\documentclass[10pt]{article}

\usepackage{algorithm}
\usepackage{algpseudocode}
\usepackage{graphicx}
\algnewcommand{\Inputs}[1]{%
  \State \textbf{Inputs:}
  \Statex \hspace*{\algorithmicindent}\parbox[t]{.5\linewidth}{\raggedright #1}
} 
\begin{document}






%

\title{A Cell-Division Search Technique for Inversion with Application to Picture-Discovery and Magnetotellurics}
%
%
%
%
%

%
\author{
%
%
Brad Alexander\\
School of Computer Science \\
University of Adelaide\\
\texttt{bradley.alexander@adelaide.edu.au}
\and
Yang-Heng Kee\\
School of Computer Science\\
University of Adelaide\\
\texttt{yang.kee@adelaide.edu.au }}

\maketitle
\begin{abstract}
Solving inverse problems in natural sciences often requires a search process to find explanatory models that match collected field data. Inverse problems are often under-determined meaning that there are many potential explanatory models for the same data. In such cases using stochastic search, through providing multiple solutions, can help characterise which model features that are most persistent and therefore likely to be real. Unfortunately, in some fields, large parameter spaces can make stochastic search intractable. In this work we improve upon previous work by defining a compact and expressive representation and search process able to describe and discover two and three dimensional spatial models. The search process takes place in stages starting with greedy search, followed by alternating stages of evolutionary search and a novel model-splitting process inspired by cell-division. We apply this framework to two problems - magnetotellurics and picture discovery. We show that our improved representation and search process is able to produce detailed models with low error residuals.
\end{abstract}

%



\section{Introduction}
Many problems in the natural sciences involve a process of finding an explanatory model that best fits a set of field data. Such problems are called inverse problems\cite{colton2012surveys,tarantola2005inverse}. Examples of inverse problems are varied and include, tomography, oceanographic sensing, gravitational and seismic sensing and magnetotellurics. Inverse problems are often {\em{under-determined}} that is there are many models that explain the field data. Because of this multiplicity of explanations there is a role for population-based stochastic search in solving inverse problems. Through the provision of a population, within in each run and by producing a different population between runs, these search algorithms provide a 
a variety of well-fitting solutions. Field experts can then query these solutions to look 
for distributions of features\cite{tarantola2005inverse}.

Unfortunately, inversion using stochastic search is computationally expensive
and can even be infeasible when models are described by many parameters. This is 
typically the case in magnetotellurics where 3D models defining the 
resistivity of the Earth's subsurface can consist of thousands of parameters.
In earlier work \cite{alexanderstuba} we 
described a methodology, called blob-modelling which reduced the number
of parameters needed to describe magnetotelluric models. This parameter reduction made stochastic search practical for 3D magnetotelluric inversion.  Later we performed a case study\cite{alexanderstubb} which showed that, while the blob-modelling method was reasonably effective in a variety of settings it was sensitive to the starting configurations for the search process. We also found later that this initial approach struggled to represent more detailed models.

This paper describes a substantially refined approach to inversion of problems involving 2D and 3D geometries. The technique involves a multi-stage search involving and initial greedy search phase, followed by interleaved stages of evolutionary search, model simplification (culling) and cellular division (blob-splitting). The cellular division phase is carried out by splitting ellipse functions that make the largest contribution to fitness. We also substantially improve the representation by adding a strength parameter to each ellipse function to smoothly express the degree of local dominance. 

These changes have improved the outcomes and reliability of previous work. The cell-growth model in particular allows for detailed models to be evolved with more success than our previous all-at-once approach.

\subsection{Contributions}
The contributions of this paper are as follows: We describe a novel approach inversion by combining evolutionary search with cellular division; We describe an improved way to express 2D and 3D models using ellipse functions with a strength parameter that denotes the degree to which an ellipse is in the foreground; We describe an improved initial search phase, based on the Hough transform\cite{mukhopadhyay2015survey} but applicable to a large number of diffuse ellipse functions of different colours.

We apply our technique to an artificial benchmark problem of discovering unknown 2D pictures. We then extend the technique back to the original problem of 3D magnetotelluric inversion. We conduct 
experiments that demonstrate the advantages of adding blob-splitting to the search process.
We show that our technique ports between the benchmark problems with very little modification which indicates promise in the technique generalising to other fields. 

\subsection{Paper Structure}
The rest of this article is laid out as follows. In section~\ref{related} we put our work in context of other work in the problem domain. In section~\ref{technique} we describe our technique as applied to both 2D and 3D problems in picture discovery and magnetotelluric inversion. In section~\ref{benchmarks} we describe the customisation of our framework to our two 
benchmark problem domains. In section~\ref{results} we present our results and, finally, section~\ref{conclusion} concludes and canvases future work. 

\section{Related Work}\label{related}
There is a long history of the application of stochastic search to inverse problems\cite{colton2012surveys}. This history is reflected in stochastic approaches to inversion in magnetotellurics (MT)\cite{Jones79,Sambridge02,Grandis02,Flores02,Schwarzbach05,Tong09,Moorkamp10,liu12,alexanderstuba}. 
The early approaches included Monte-Carlo 
techniques\cite{Jones79,Sambridge02}, which provide better global search at the 
cost extra computation arising from random search. Faster search is provided
by informed
search heuristics such as Markov chaining\cite{Grandis02}, 
simulated annealing\cite{Dosso91}
and evolutionary methods\cite{Flores02,Schwarzbach05,Tong09,Moorkamp10,liu12}. 

Work using evolutionary 
methods is diverse, ranging from direct evolution of 2D models
using specialised operators (Flores and Schultz\cite{Flores02});
to hybrid schemes combining genetic algorithms and local search (Tong et. al.\cite{Tong09}); to using pareto-optimisation techniques to balance model fitness
and model smoothness (Moorkamp, Jones, and Fishwick\cite{Moorkamp10}). Work in 3D inversion
has been limited due to the large number of parameters involved, Liu and Li, et. al.\cite{liu12} inverted 3D models 3600 parameters using a real-valued GA with informed mutation operators
and inequality constraints to prevent divergence. (Anonymised..)\cite{alexanderstuba,alexanderstubb} used reduced-parameter representations and model priming to invert 3D models. This latter work forms the foundations for the work described here.
Some work has also been done on using reduced-parameter representations for 
MT\cite{Jones79,Schnegg99} our work differs in the number of separate elements and 
the lack of assumptions about the prior knowledge of the model. 

In relation to picture discovery, because our benchmark was deliberately set up to 
take no advantage of information beyond a scalar error value,
 there are few direct analogues to this benchmark in the literature on signal processing.
However there are works that share features. 
The scanning step in the priming stage is similar
to some implementations of the Hough transform\cite{mukhopadhyay2015survey}. 
Our work is also related
to work in circle detection and image segmentation\cite{ayala2006circle,cuevas2013comparison}. More recently, recently Cuevas et. al.\cite{cuevas2014} used a competitive evolution process, based on animal behaviour to detect ellipses in an image.

Our work differs in that it expresses a whole picture in terms of ellipses rather than detecting specific elements within pictures. In addition our work, doesn't exploit priors, which makes it much slower for applications where detailed prior information is available but much more applicable to situations where prior knowledge is limited.   

\section{Methodology and Background}\label{technique}
The aim of our method is to perform 2D and 3D spatial inversion. 
The search process cyclically refines guesses of the model in response to error values. 
An inverse problem can be formulated as\cite{colton2012surveys}: $F(m)=d$ where $d$ is
the data signal (responses) 
collected and $m$ is the model being sought. The function $F$ is the 
{\em{forward}} function which maps a model to its data signal. Given $F$ and $d$ it
is possible to derive $m$ through a process of estimation and refinement as outlined in 
Algorithm~\ref{process}. 
\begin{algorithm}[t]
\caption{General Algorithm for Inversion}\label{process}
\begin{algorithmic}[1]
\Procedure{Inversion}{$\boldmath{d}$}\Comment{find model fitting $\boldmath{d}$}
\State initialise candidate models $m$
\Repeat
\State $\boldmath{m} \gets \mbox{\em{improve}(m)}$ \Comment{improve the models}
\State $\boldmath{r} \gets \mbox{F}(m)$ \Comment{get model responses}
\State $\mbox{\em{errs}} \gets | \boldmath{r}-\boldmath{d} |$ \Comment{calculate model errors}
\Until{ $\mbox{\em min}(\mbox{\em{errs}}) < \epsilon$}
\State {\bf{return}} $\mbox{\em best\_of}(m)$
\EndProcedure
\end{algorithmic}
\end{algorithm}
The input to the process is a field data vector $\boldmath{d}$. The output is the best models produced by the process.  The process first initialises $m$. The process then iterates through successive steps of improvement and testing. The {\em{improve}} function perturbs the models $m$ through either random or informed search heuristics and can vary during the search process. The $F$ function takes the models $\boldmath{m}$ and uses it to produce simulated responses $r$ that can then be compared to field data. This {\em{forward-modelling}} step is key to the inversion process and is different for each problem domain.
The loop terminates when the minimum error $\mbox{\em{errs}}$ falls below a threshold, at which point the best models are returned.

In this work our two benchmarks have a different forward-modelling function. For the picture discovery (PD) problem our forward model samples our 2D model representation into a 100-by-100 grayscale picture. For magnetotellurics (MT) our forward model samples our 3D model representation to a hexahedral mesh. This mesh is then given to an MT forward modelling function written by Siripuvaraporn\cite{Siri12} which generates a simulated field response for comparison to the real field response. 

We describe our model representation, evaluative function, and our search process next. 
In our discussion we describe the process for the PD problem and highlight any modifications required for the MT problem. We also highlight the differences in methodology from our previous published work. 
\subsection{Representation}
Our representation needs to be compact enough to make stochastic search feasible but expressive enough to admit realistic models. We represent our model as
a background colour: $b$ combined with a set of overlapping diffuse ellipse functions $f_i$. 
\[
  (b,\{f_1,\ldots,f_n\})
\]
The background value $b$ is new to this work and allows the models background colour to be determined as part of the search process\footnote{We found that including this term helped search performance in our early experiments.}.

The ellipse functions (blobs) are parameterised by central intensity, strength, diffuseness,  position, size, and rotation. These parameters are
described in Table~\ref{blobparams}. All parameters are normalised 
to the range $[0,1]$ and represent values pertinent to their purpose. Thus
a value of $0.0$ for $x_{loc}$ will place the centre of the blob on the left 
side of the image/model; a value of $1.0$ for $z_r$ will rotate the blob by $\pi$\footnote{In theory, because ellipses are symmetrical, we could limit rotations to $\pi/2$ without loss of generality but this can make search difficult for values just beyond that limit.}; and a value of $0.5$ for $x_r$ will give the blob an x-radius of half the size of the model.  
\begin{table}
\begin{center}
\begin{tabular}{|l|p{4.5cm}|}
\hline
Param & Description  \\ \hline \hline
$\delta$ & central intensity \\ \hline
$s$ & strength \\\hline
$\alpha$ & sharpness  \\\hline
$x_{loc},y_{loc}$ & x,y-location  \\\hline
$x_s,y_s$ & x,y \\\hline
$z_r$ & rotation about z axis \\\hline
\end{tabular}
\end{center}
\caption{Summary of blob parameters}\label{blobparams}
\end{table}
For the 3D ellipses used for the MT problem we add parameters: $z_{loc},z_s$ for z-location and size and $x_r,y_r$ for rotation about the x and y axes of the ellipse. 

The strength parameter is new to this
work and defines the extent to which the blob appears in the foreground
or background.  

The value at a location in a picture or model element is determined by both 
the local intensities the blob functions that overlap at that location and the relative {\em strengths} of the blobs at that location.

The local intensity of a blob function $f_i$ at location $(x,y)$ is given by:
\[
   f_i(x,y)= \delta_i / ((x_{tr}^2 + y_{tr}^2)^{15 \alpha} +1)
\]
The parameters $x_{tr}$ and $y_{tr}$ are the transformed values of the $(x,y)$
parameters after taking account of translation induced by the $x_{loc},y_{loc}$ 
and rotation induced by $z_r$. The numerator $\delta_i$ is the peak intensity
of $f_i$. The $\alpha$ parameter determines how sharply this intensity trails off. An $\alpha$ value close to one produces a very sharp edge while a value close to zero produces a very diffuse ellipse. Generalisation to three dimensions
entails adding a $z_{tr}$ parameter to the above equation.

At each point of the model the local intensities of each blob are combined to produce a value for that location. In previous work we combined intensities using a generalised mean but this method tended to smear features of moderate intensity. In this work we use the strength parameter $s_i$ determine the extent to which blob $i$ is locally dominant. Local dominance can be viewed as the extent
to which the blob is in the foreground. 

\begin{figure}
\begin{center}
\includegraphics[width=4cm]{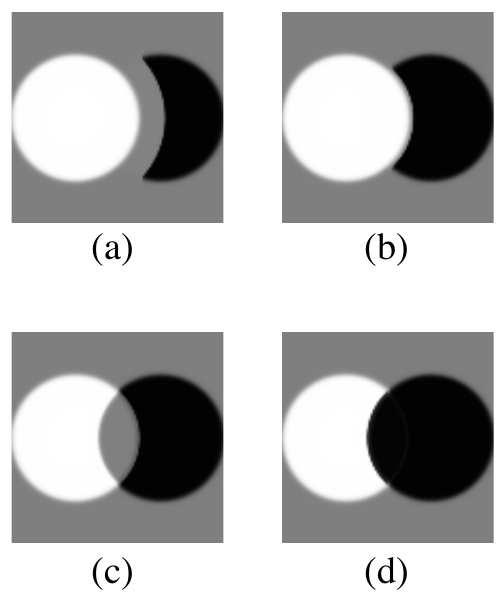}
\end{center}
\caption{Illustration of effect of strength parameter. Left and right blobs have
respectively (a): $s=1,0$, (b) $s=0.7,0.3$, (c) $s=0.5,0.5$, and (d) $s=0.3,0.7$. }\label{localdominance}
\end{figure}

To create a combined model value we first
 assign the background intensity to a dummy blob intensity $f_o(x,y)$
and set $s_0=0$ (background strength is zero). Then for the remaining blobs we 
normalise the local intensity $f_i(x,y)$ with respect to the maximum intensity
of that blob to give $f'_i(x,y)$. We also normalise strengths with respect to 
the maximum strength blob. To help make foreground and background blobs more distinct we raise this normalised strength to the sixth power. We denote these adjusted strengths: $s'_i$. From these values for $f'_i(x,y)$ and $s'_i$ the combined local value $v_{(x,y)}$ is produced from equation~\ref{eq:v}
\begin{equation}\label{eq:v}
v_{(x,y)}=\mbox{\em{bg}}+\mbox{\em{fg}}\times(1-(\mbox{\em{bg}}/\mbox{\em{wi}}))
\end{equation}
where {\em{bg}} is defined: $\mbox{\em bg}=\sum_{i} (1-s'_i)\times f'_i(x,y)$ and represents the
extent to which the model background influences this location. It's dual, {\em{fg}} is defined: $\mbox{\em fg}=\sum_{i} s'_i\times f'_i(x,y)$ denotes the extent to which the foreground influences this location. The denominator {\em{wi}} is defined: $\mbox{\em wi}=\sum_{i} s'_i\times\delta_i$ is a weighted intensity for the whole model. $wi$ is used preserve the influence of the background term when the overall model is low intensity at the given location. 
Equation~\ref{eq:v} works to allow relatively high strength blobs to dominate the value at each location. Conversely if location $(x,y)$ is overlapped only by low strength blobs then the background term will dominate. It should be noted that extending Equation~\ref{eq:v} in the third dimension trivially involves the addition of a $z$ parameter. 
The effect of the strength parameter blob influences is illustrated in Fig~\ref{localdominance}. 

Finally,  it should be noted that the value of $v_{(x,y)}$ is in the 
range $[0,1]$ this value has to be normalised to each application as described in section~\ref{benchmarks}.
\subsection{Evaluative Function}
Our framework evaluates each individual by measuring the difference between the response 
of the individual $r$ and the input data for the inversion $d$. For an individual evaluation this difference is 
represented by a single scalar error value. This is the only value used to guide the 
search process.
In this work we do not attempt to add a regularlisation term to our error value. This is 
primarily
because the use of blobs keeps models relatively simple.
The details of the extraction of the response value for our picture-discovery and magnetotellurics benchmarks are described in section~\ref{benchmarks}. 

\subsection{Search Process}
The search process is defined by the algorithm shown in Algorithm~\ref{search}.
\begin{algorithm}[t]
\caption{Global Search Process}\label{search}
\begin{algorithmic}[1]
\Procedure{Search}{$\boldmath{d}$}\Comment{find model fitting $\boldmath{d}$}
\State initialise blank model $m$
\State $m \gets \mbox{\em prime}(m,d)$  \Comment{greedy search}
\State $\mbox{\em{pop}} \gets \mbox{\em{cma-es}}(m)$  \Comment{run CMA-ES}
\State $m \gets \mbox{\em best}(\mbox{\em pop})$
\For{ $i$ in $1$ to {\em num\_iters}}
\State $m \gets \mbox{\em cull}(m)$   \Comment{remove bad blobs}
\State $m \gets \mbox{\em split}(m)$  \Comment{split influential blobs}
\State $\mbox{\em{pop}} \gets \mbox{\em{cma-es}}(m)$  \Comment{run CMA-ES}
\State $m \gets \mbox{\em best}(\mbox{\em pop})$
\EndFor
\State {\bf{return}} $m$
\EndProcedure
\end{algorithmic}
\end{algorithm}
The first step in the algorithm is {\em{prime}}. This greedy function
takes a blank model and adds and adjusts blobs until no further 
improvements can be made. Each blob is added by scanning a diffuse light blob and then a diffuse dark blob across the model until the position returning the highest fitness for each is found. The position, shape, orientation and intensity is then refined by half-interval search until no further improvements can be made.
The priming function then chooses the refined dark or light blob according to 
whichever one contributes most to model fitness and discards the other. The background is then adjusted to improve fitness. This process of adding blobs is then until no further improvement in fitness is possible. 

It should be noted that this initial, scanning, stage of blob addition is similar to the process for the Hough transform\cite{mukhopadhyay2015survey}. However, priming differs in that the search for the non-positional parameters is greedy and non-exhaustive. The priming process also differs from our earlier work~\cite{alexanderstuba} which started with a user-defined number of blobs in set locations before proceeding straight to half interval search. In this previous work it was challenging to 
profitably place more than 16 blobs where as the current priming process, 
depending on target model complexity, can routinely place more than 30 blobs. 

\begin{figure}
\begin{center}
\includegraphics[width=5cm]{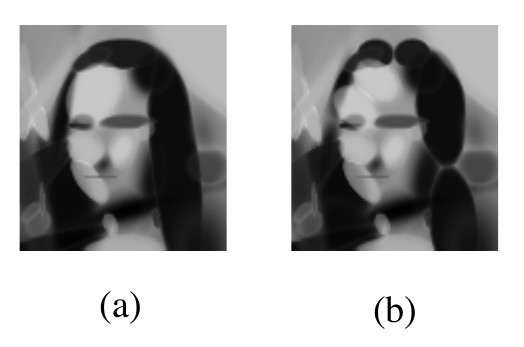}
\end{center}
\caption{Illustration of splitting. A picture model before splitting (a) and after the most significant three blobs are split (b). Note that splitting induces
a temporary decrease in model fitness.}\label{split_demo}
\end{figure}

After priming,  Algorithm~\ref{search} invokes the Covariance Matrix Adaptation - Evolutionary Strategy (CMA-ES)\cite{Hansen03} to optimise the parameters of the best model from the priming stage.
After CMA-ES the best value from the population is extracted and a loop is entered. The first operation in the loop is {\em{culling}}. This greedy operation test the contribution of each blob and if its contribution is negative, it is deleted. The next operation is splitting which divides a user-defined number (usually 3-5) of the most significant blobs into two. The division takes place along the longest axis of the original blob and the resulting blobs are 2/3 the radius of the original blob and shifted along the long axis so they just touch. The effect 
of splitting on a picture discovery benchmark image  is shown in figure~\ref{split_demo}.
The aim of splitting is two-fold, first it allows significant elements to become further refined, second it disrupts the model allowing a new phase of exploration. After splitting the model is further evolved and culled in later iterations until a set limit on the number of splits is reached.  

This concludes the description of the search algorithm. Next we describe
the setup for the benchmarks for our experiments.

\section{Benchmark Setup}\label{benchmarks}
We use two types of benchmarks in our experiments: picture-discovery (PD) and 
magnetotellurics (MT). For picture discovery we use the target images shown in Fig~\ref{picture_results}(a). For PD our evaluative function uses the function in 
Equation~\ref{eq:v} to sample its input genome at each position of a 100x100 grayscale image. These samples are then scaled to discrete integers in the 
range 0-255 to create pixel values in a grayscale picture. The error is calculated by 
subtracting the target picture from the candidate picutre pixel-wise and averaging the 
absolute values of the result. 

For the MT benchmarks we aim to model the resistivity of a 3D segment of planetary crust by finding a model to match the response of ground-based field stations to naturally occuring eletromagnetic radiation. To do this we sample the genome into each element of a 3D hexahedral mesh. The samples $v_{x,y,z}\in[0,1]$ are mapped into resistivity $\mbox{\em res}_{(x,y,z)}$ by the following equation. 
\[
\mbox{\em res}_{(x,y,z)} = 10^{-2+5v_{x,y,z}}\Omega m 
\]
which allows the model to express relatively
shallow, conductive earth with reasonable 
fidelity. To evaluate the error for a model the sampled model is run through a forward
modelling function 
using Siripuvaraporn's wsinv3dmt\cite{Siri12} to produce an artificial response
that can then be subtracted from the field response to obtain an RMS error value. 

Our first MT target model is the artificial COMMEMI 3d2 model\cite{Zhdanov97} pictured in Fig~\ref{comm:models}(a). This model is one of a series of models developed to test MT inversion techniques. It is a reasonably challenging target in that it has highly contrasting bodies near the surface and alternating conductive and resistive layers. We use a relatively low resolution model of 
13x14x10 cells to keep the evaluation time for each model to less than 2 seconds. To create artificial field data from the model we use wsinv3dmt to run a forward model to convert the model parameters into impedance tensors for five frequencies (2Hz to 0.01Hz) measured by a set of simulated field stations positioned over the model. 

Our second MT benchmark is real field data from a site near Paralana in South Australia that shows promise as an enhanced geothermal power source\cite{peacock2013time}. The response data was collected from 54 stations and processed tensors in six frequencies ranging from 50Hz to 0.01Hz. For the inversion we again, use a low-resolution model (13x13x10 cells).

\section{Results}\label{results}
We conducted three experiments (one picture discovery and two MT). Our first experiment was applied to the 
PD benchmark, with splitting (PDSplit) and without splitting (PDNoSplit). 
This experiment was used to 
verify that the framework could be used to model complex environments and 
that splitting is potentially profitable. 

Our second experiment was applied the discovery the COMMEMI 3d2 MT benchmark.
In this experiment we compare the results of multiple runs using splitting (MTSplit), no splitting (MTNoSplit) and a single run the standard (deterministic)  wsinv3dmt gradient search program (MTWsinv).

Our third experiment (ParaSplit) completes several runs of the 
configuration used in 
MTSplit on the Paralana data set. We compare the runs to each other and 
to the the outcome of wsinv3dmt (ParaWsinv) applied to the data set. 

All experiments were conducted on a 48processor Intel i7 platform with 64GB of memory. In all cases we configured CMA-ES with $\mu=50$,$\lambda=25$ and $\sigma=0.01$. We describe 
each experiment in turn. 

\subsection{Picture Discovery}
\begin{figure}
\begin{center}
\includegraphics[width=6cm]{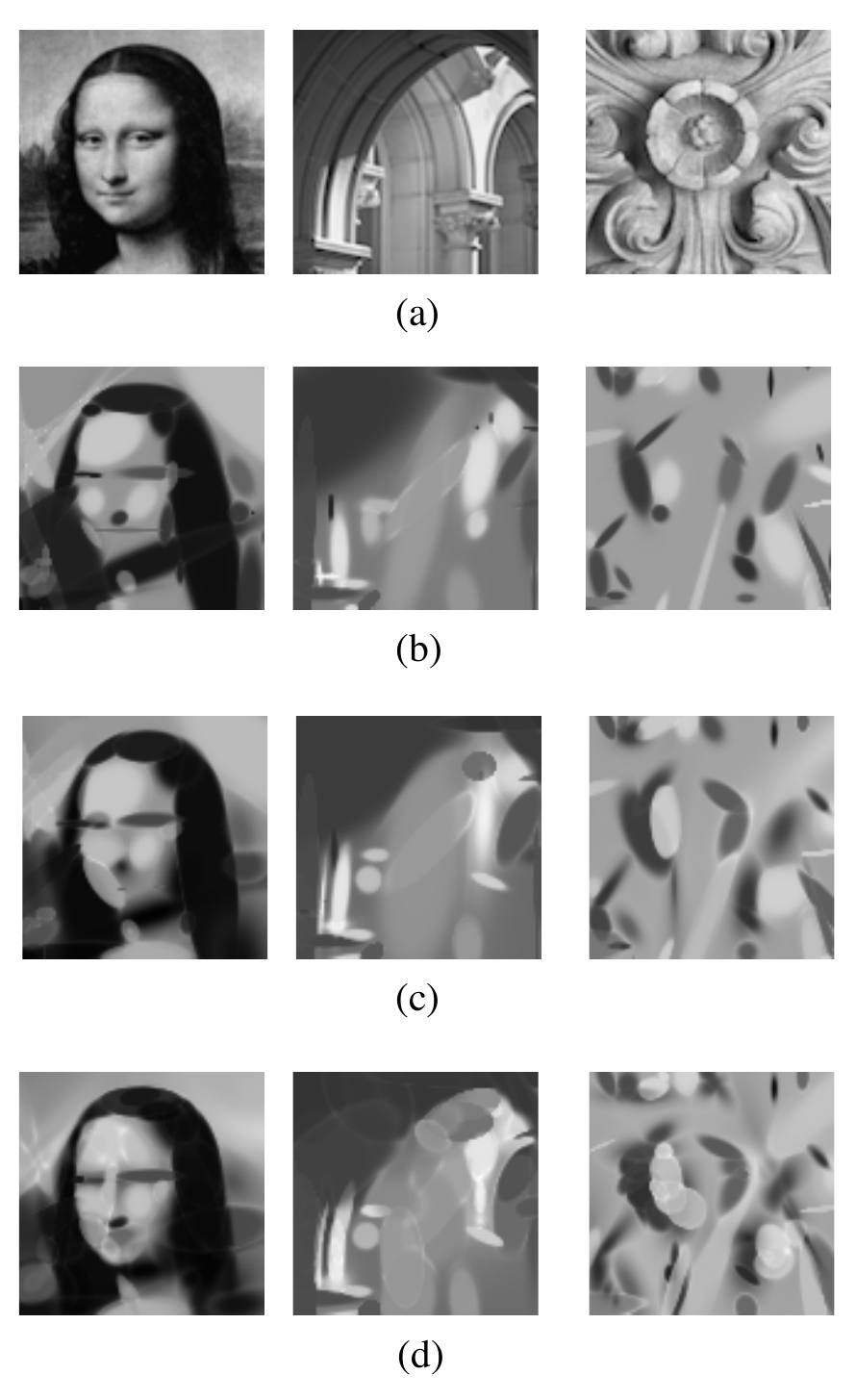}
\end{center}
\caption{Results of PD experiments. (a) the target images. (b) pictures after priming. (c) results from PDNoSplit (d) results from PDSplit.}\label{picture_results}
\end{figure}

In this experiment we ran a splitting (PDSplit) and non-splitting (PDNoSplit)
version of our search algorithm once on each of the PD benchmarks pictured in Figure~\ref{picture_results}(a). PDSplit ran with five splits of up to five blobs each interspersed with six runs of CMA-ES with a maximum of 10000 iterations (producing a maximum of 3000000 evaluations in the CMA-ES stage). The PDNoSplit algorithm ran with the same priming as PDSplit and one stage of CMA-ES with a maximum of 60000 iterations. 
The results show that PDSplit (part (d)) performs visibly better on all three benchmarks
than PDNoSplit (part (c)).
The error values for PDSplit are also smaller by 26\%,17\% and 12\% respectively on the 
the three benchmarks. However, PDNoSplit terminated
after only 6000 of its 60000 iteration due to lack of progress. This means
that PDSplit could be benefiting from both the additional model parameters
produced by the splitting and the additional evaluations enabled restarting
CMA-ES after splitting. As a result PDSplit was able to run between 6000 and 9000 iterations
in each of its CMA-ES phases before termination for flat-fitness. In total each run for the
PDSplit benchmark took approximately 3 days with 98\% of the processing time devoted to 
CMA-ES.

Finally, it should be noted that a 2d version of 
our previous framework\cite{alexanderstuba} (implemented early in the refinement process)
was not able to make significant progress in 
the priming stage of the benchmarks.

\subsection{Magnetotellurics: COMMEMI 3d2}
\begin{figure}
\begin{center}
\includegraphics[width=6cm]{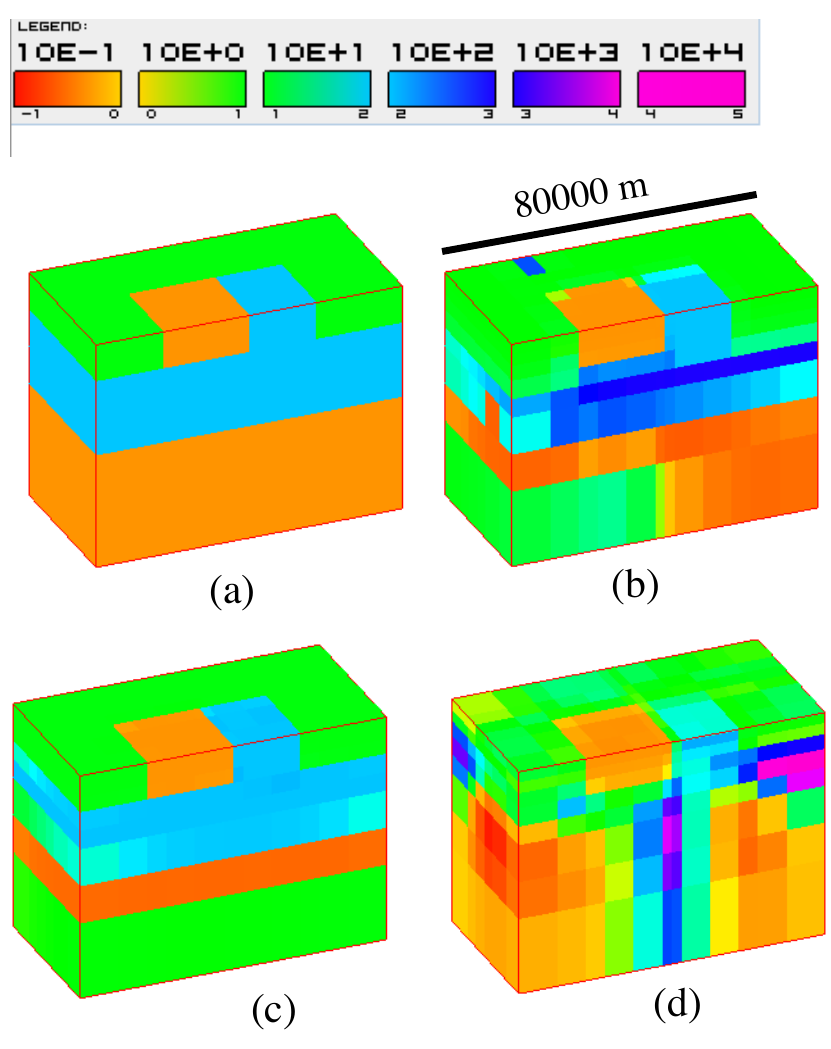} 
\end{center}
\caption{Cross sections of: (a) the COMMEMI 3d2 target model, (b) the best
model produced by MTNoSplit, (c) the best model produced by MTSplit, (d) the 
best model produced by wsinv3dmt. }\label{comm:models}
\end{figure}

For the MT experiments we used the same model setup and number of evaluations as for our previous proof-of-concept work\cite{alexanderstuba} to allow for a comparison with the fittest values produced for that work.  The target model is the COMMEMI 3d2 model\cite{Zhdanov97} pictured in Fig~\ref{comm:models}(a).

We ran two trials of 20 and 17 runs respectively\footnote{Three of the runs in the second experiment were aborted due to system limits.}. In the first trial (MTSplit) we primed the model to 4 blobs and then, ran CMA-ES four times with 1000 iterations, interspersed with splitting and culling to bring the model to 13 blobs. In the second trial we primed to 13 blobs and ran CMA-ES once with 4000 iterations - thus equalising the number of blobs and the number of iterations. For reference also ran a deterministic gradient-search inversion using wsinv3dmt  for 50 iterations. In all experiments the starting background was set to $10\Omega m$ which is the background resistivity at the surface of the target model. 

The results of the experiment are summarised in Table~\ref{mt_res}.
\begin{table}
\begin{center}
\begin{tabular}{|l|l|l|l|l|}
\hline
Name & Runs & Best & Worst & Mean  \\ \hline \hline
MTNoSplit & 17 & 0.380 & 0.743 & 0.486 \\\hline
MTSplit & 20  & 0.102 & 0.306 &  {\bf{0.156}}\\ \hline
wsinv3dmt & 1 & 1.678 & 1.678 & 1.678  \\\hline
\end{tabular}
\end{center}
\caption{Summary of results from the COMMEMI experiments}\label{mt_res}
\end{table}
MTSplit performed significantly better than MTNoSplit with 
$p < 0.001$ on the Wilcoxon sum rank test. 
Note again that wsinv3dmt, from a given starting model, 
is deterministic, so all runs yield the same
results. The best value for wsinv3dmt of $1.678$ was achieved in the 31st iteration. 

Fig~\ref{comm:models} shows cross-sections for the target model (part (a)) and the best models produced by MTNoSplit, MTSplit and wsinv3dmt respectively.
As can be seen, MTSplit appears to produce the best approximation to the target
model, especially at the surface where signals are most distinct. The conductive layer at depth is thinner than in the target but, given the relative opacity of conductive layers in MT these models will appear nearly identical 
to the surface
sensors\footnote{This is, in part, an artifact of MT inversion being underdetermined - MT is analogous to tomography carried out from one side of the target object.}. It should be noted that the performance of wsinv3dmt, being a gradient search method, will depend on starting model as well as model resolution so further experiments are needed for a fully valid comparison with the other methods. It should also be noted that wsinv3dmt is much faster than the stochastic methods, taking 2 hours of runtime in contrast to 4 days for MTNoSplit and MTSplit. As an additional note both MTSplit and MTNoSplit outperformed the runs described in our earlier work\cite{alexanderstuba} which produced an RMS of 1.28. 

\begin{figure}
\begin{center}
\includegraphics[width=8cm]{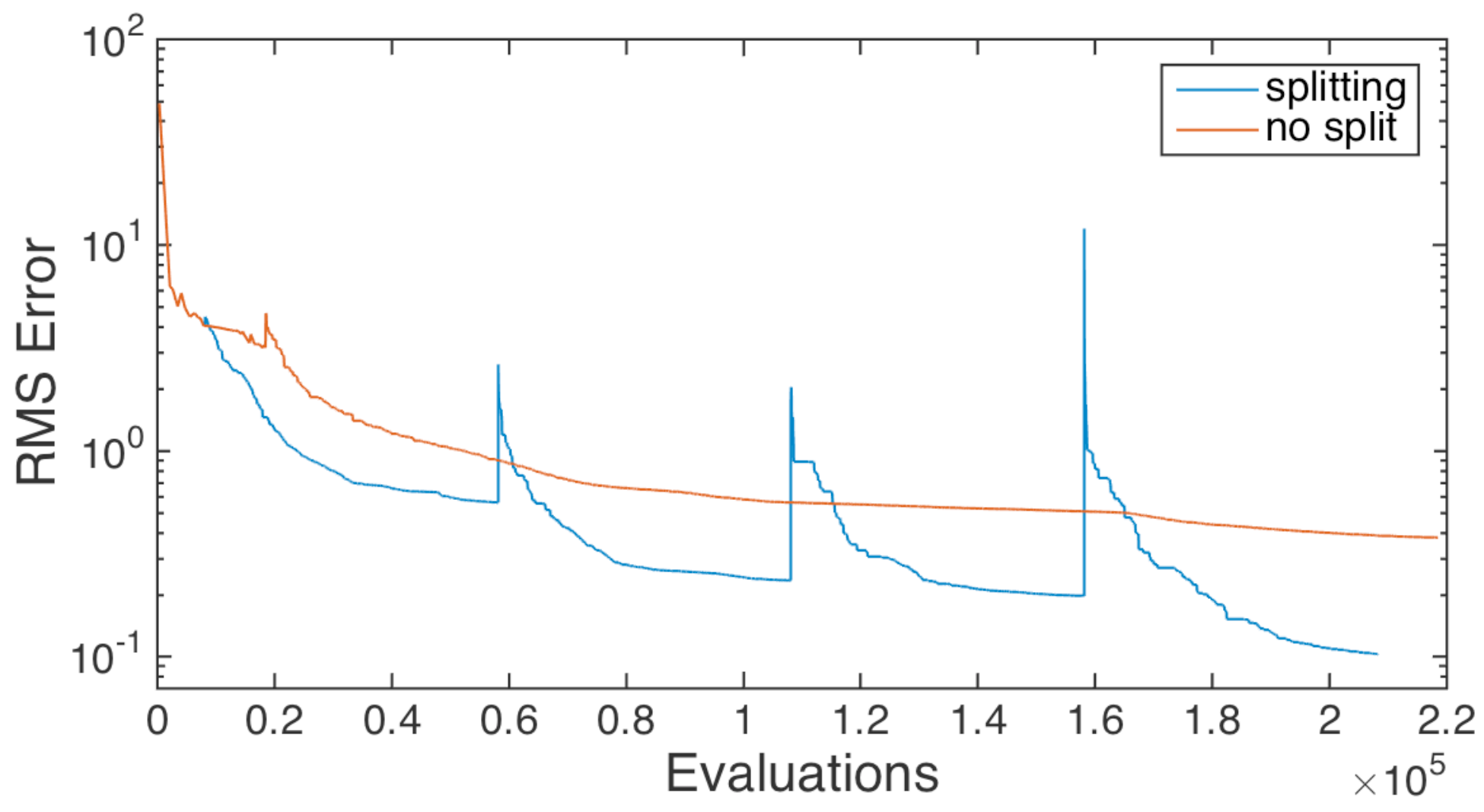}
\end{center}
\caption{Log plot of error values for MTNoSplit (red line) and MTSplit (lower blue line). MTSplit adds blobs incrementally by splitting leading to better and slightly faster convergence.}\label{graph_plot}
\end{figure}

In order to contrast the evolutionary processes followed by MTNoSplit and MTSplit we plot the log of error values against evaluations in Fig~\ref{graph_plot}.
The figure shows the best MTSplit run exhibits improved performance through
faster evolution of simpler models at the start (4 blobs versus 13 blobs) and
through the restarting of evolution after the splits which are evident as
peaks in the error value. MTNoSplit also takes slightly longer due to a longer
priming stage (approximately 19000 evaluations vs 8000 evaluations for MTNoSplit).

\subsection{Magnetotellurics: Paralana}
\begin{figure}
\begin{center}
\includegraphics[width=6cm]{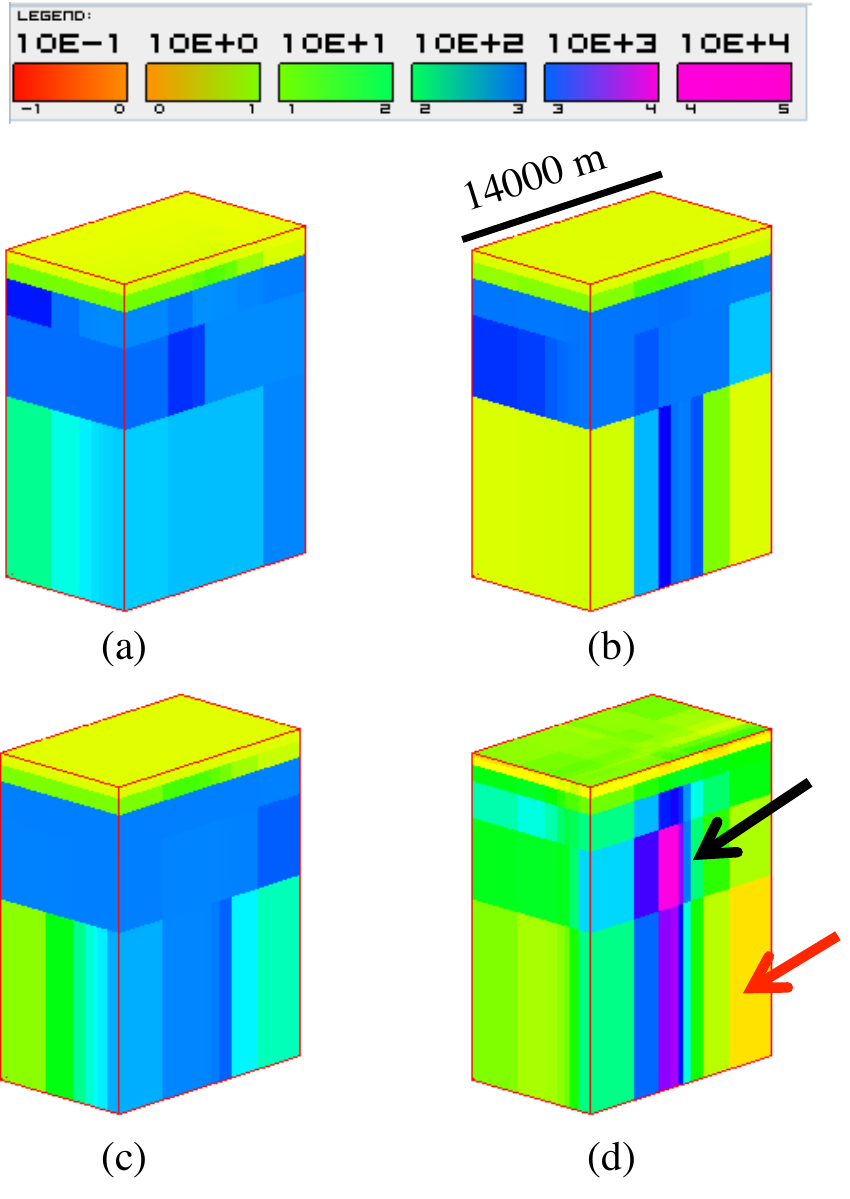}
\end{center}
\caption{The first (a), second (b) and third (c) best models from ParaSplit 
and the model from ParaWsinv (d). The black arrow denotes a known area 
of hot dry rocks. The red arrow denotes a posited heat source.}\label{para}
\end{figure}

As with the COMMEMI 3d2 model, we used the same model parameters and data
as for our previous work. In this experiment we conducted 5 trial runs with 
splitting (ParaSplit) and one reference run of 50 iterations with wsinv3dmt (ParaWsinv). Each run of ParaSplit primed to 4 blobs and interspersed 5 runs of CMA-ES (1000 iterations)  with 4 rounds of splitting and culling  with each round adding 4 blobs\footnote{Culling removed approximately two blobs in each run}. The runs for ParaSplit each took 5 days to complete. The RMS's of the ParaSplit runs ranged between 2.440 and 2.556. Four out of five of these runs are
slightly better than the run of 2.51 from our previous experiments. The ParaWsinv run performed better than ParaSplit with a minimum RMS of 2.18 in its third iteration. Figure~\ref{para} shows the models produced by the best three runs of ParaSplit and ParaWsinv.
respectively. 
All four models in the figure exhibit some common structure with a relatively conductive surface layer underpinned by a thick region of less conductive material with indication of more conductivity at depth. However, the three ParaSplit models\footnote{The remaining two ParaSplit models are very similar in structure to the three models displayed.} have much more extensive regions 
moderate resistivity compared to ParaWsinv which has a smaller region of much 
higher resistivity ($> 10000\Omega m$) close to known structures of hot-dry rocks (marked with a black arrow in the image). ParaWsinv also has a strongly defined area of higher 
conductivity at depth (marked with a red arrow). This corresponds with a posited heat source arising from radioactive decay in host rocks. One possible explanation for these differences is that the experimental setup for ParaSplit limited
resitivty to a maximum of $1000 \Omega m$ which prevents its models expressing
the higher levels of resistivity shown in part (d). The larger areas of moderate resistivity in parts (a) through to (c) may be compensating for an inability to express higher resistivity but confirmation of this 
will require further experimentation. 

\section{Conclusions and Future Work}\label{conclusion}
In this paper we described a new methodology for refining spatial models
during inversion. We have demonstrated that this methodology can enhance 
search and enable the building of of more detailed 2D models in picture-discovery benchmarks. We applied our methodology to a popular
3D magnetotelluric benchmark and 
demonstrated significant advantages in search compared to non-splitting search and to our earlier published results. We also showed that our methodology can be applied to real 3D models - but perhaps requires some caution in setting of search space bounds for resistivity. 

This work can be extended in several ways. First, for problems - such as 
picture discovery - where significant prior knowledge is available - exploitation of that knowledge can be built into the priming stage of that search. Second
the technique should be applied to higher resolution models to verify that 
its performance is preserved as sampling intervals are decreased. Finally,
given the relatively small number of parameters in our models it may be 
possible to automatically learn a surrogate magnetotelluric forward modelling function 
- at least for very simple models. Such a function could potentially speed up the intial stages of inversion by two or three orders of magnitude. 
\section{Acknowledgments}
Thank you to Stefan Thiel and Jared Peacock
for their support and advice in the development of the refinements described in this paper.
Thank you to Petratherm Pty Ltd for access to the Paralana modelling data. 

%
\bibliographystyle{abbrv}
\bibliography{evol_mt}  
%
%
\end{document}